\pdfoutput=1

\documentclass[11pt]{article}

\usepackage{emnlp2021}

\usepackage{times}
\usepackage{latexsym}

\usepackage{microtype}
\usepackage{graphicx,epsfig,subfig,diagbox}
\usepackage{algorithm}
\usepackage{algorithmic}
\usepackage{hyperref}
\usepackage{multirow}
\usepackage{rotating}
\usepackage{url}
\usepackage{amsmath,graphicx}
\usepackage{caption}

\usepackage[T1]{fontenc}

\usepackage[utf8]{inputenc}

\usepackage{microtype}

\title{On Building Spoken Language Understanding Systems for Low Resourced Languages}

\author{Akshat Gupta \\
  J.P.Morgan AI Research \\
  New York, USA \\
  \texttt{akshat.x.gupta@jpmorgan.com}}

\begin{document}
\maketitle
\begin{abstract}
Spoken dialog systems are slowly becoming and integral part of the human experience due to their various advantages over textual interfaces. Spoken language understanding (SLU) systems are fundamental building blocks of spoken dialog systems. But creating SLU systems for low resourced languages is still a challenge. In a large number of low resourced language, we don't have access to enough data to build automatic speech recognition (ASR) technologies, which are fundamental to any SLU system. Also, ASR based SLU systems do not generalize to unwritten languages. In this paper, we present a series of experiments to explore extremely low-resourced settings where we perform intent classification with systems trained on as low as one data-point per intent and with only one speaker in the dataset. We also work in a low-resourced setting where we do not use language specific ASR systems to transcribe input speech, which compounds the challenge of building SLU systems to simulate a true low-resourced setting. We test our system on Belgian Dutch (Flemish) and English and find that using phonetic transcriptions to make intent classification systems in such low-resourced setting performs significantly better than using speech features. Specifically, when using a phonetic transcription based system over a feature based system, we see average improvements of 12.37\% and 13.08\% for binary and four-class classification problems respectively, when averaged over 49 different experimental settings.

\end{abstract}





\section{Introduction}
\begin{figure*}[ht]
  \centering
    \includegraphics[width=0.80\linewidth]{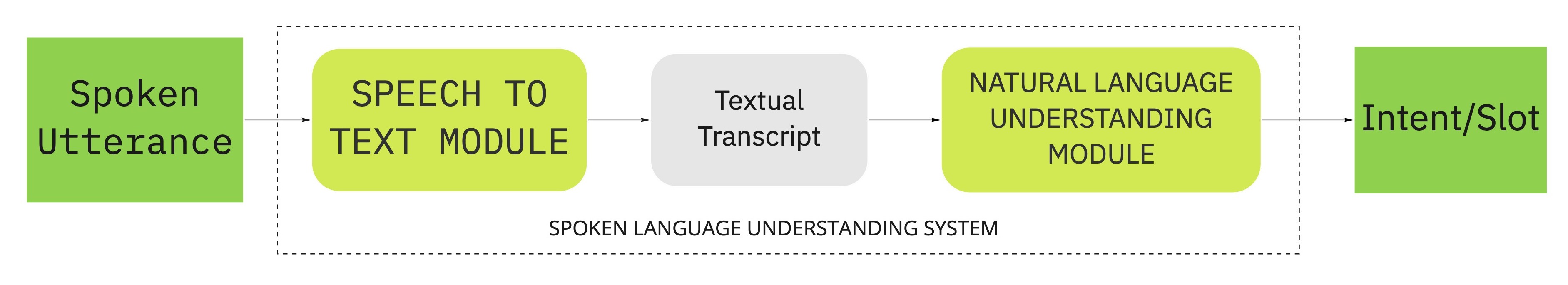}
    \caption{A traditional spoken language understanding system consisting of a speech-to-text system followed by a natural language understanding module.}\label{fig:1}
\end{figure*}

Spoken Language Understanding (SLU) systems form an integral part of any spoken dialog system. A traditional SLU pipeline is made up of two modules (Figure \ref{fig:1}) - a speech to text module which converts input audio into textual transcripts, and a natural language understanding (NLU) module which aims to understand the semantic content in the user utterance from the textual transcripts \citep{tur2011spoken, lugosch2019speech}. The conventional two-module SLU pipeline is prone to making speech recognition errors which propagate through the system. To minimize these errors, a lot of recent research has been focused on creating end-to-end spoken language understanding (E2E-SLU) systems \citep{qian2017exploring, serdyuk2018towards}. 

Building E2E-SLU systems requires an even larger amount of task-specific annotated data when compared to the two-module split SLU pipelines \citep{lugosch2019speech, bastianelli-etal-2020-slurp, wu2020harpervalleybank}. While high resourced languages like English are moving towards E2E-SLU, the challenges presented by low resourced languages are very different. Low resourced languages operate in a regime where we have access to only tens or hundreds of labelled utterances, which are not enough to build robust E2E-SLU systems. Creating robust automatic speech recognition (ASR) systems for low resourced languages is itself a challenge as these require large amounts of manual annotation. For many low resourced languages, we might not even have ASR technologies. Creating ASR technologies for unwritten languages or languages that have only a few hundred or a few thousand speakers alive is not even a viable option. But can we create spoken dialog systems for such languages? 

`\textit{Low-resourced-ness}' of a particular language is a very broad term often used loosely to describe various types of inadequacies when creating language technologies. It affects creating speech technologies in mainly two ways. For the purpose of this paper, we explicitly define and differentiate between these two scenarios. The first scenario is what we call \textit{language-specific low-resourced-ness}, where we do not have enough resources to create robust, language specific speech recognition technologies. Speech recognition systems are fundamental to creating various kinds of speech technologies including dialog systems, speech emotion recognition systems, keyword spotting systems, speaker recognition and diarization systems. When creating dialog systems, ASR systems allow us to convert input speech to text, after which text based language models like BERT \cite{devlin2018bert} can be used to understand the content of speech and build NLU modules. This allows us to create SLU systems with smaller amounts of task-specific annotated data. But in settings where we do not have access to speech recognition systems, it becomes important to have enough annotated task-specific data to compensate for the lack of ASR systems and text-based language models. This introduces the second source of `low-resourced-ness', which we call \textit{task-specific low-resourced-ness} - where we do not have enough annotated data for a particular task. Two challenges occur in this scenario - one where we do not have enough speakers to create a task-sepcific speech corpus, and another where we do not have enough recordings per speaker. Not having enough annotated data for a particular task, when combined with lack of speech recognition technologies compounds the problem of creating speech technologies for such languages. We work in this compounded low-resource setting, where we assume language specific and task-specific low-resourced-ness.

In this paper, we present a series of experiments to empirically re-create language-specific and task-specific low-resourced-ness scenarios and work in the compounded setting where we tackle both challenges at the same time. As we assume language specific low-resourced-ness, we work in a setting where we don't have access to language specific ASR systems. One way to tackle this setting is to use an ASR system built for a higher resourced language and use the transcriptions generated to perform downstream tasks as used in \citep{buddhika2018domain, karunanayake2019transfer, karunanayake2019sinhala}. It was later shown in \citep{gupta2021intent, yadav2021intent} that using language and speaker independent systems trained on many languages to extract speech features works much better than using ASR systems built for a different language, as a different language usually contains a different set of phonemes with a different phone to phoneme set mapping. When this setting is compounded by task-specific low-resourced-ness, we are at an extremely low resourced setting where each data point becomes valuable. To simulate this setting, we pose an I-class intent classification problem ($I=2,4$) where we have a varying number speakers (S) available for recording training data. Each speaker provides only k-utterances per intent for training. In this k-shot setting, we evaluate our system in a granular manner for very small values of $S$ and $k$. Specifically, we evaluate our system for $S = 1,2,3,4,5,6,7$ number of speakers,  where each speaker records $k = 1,2,3,4,5,6,7$ utterances per intent. We evaluate our SLU system on robust test sets containing hundreds of utterances collected from multiple speakers which are not present in the training set. 

We find that using language independent or multilingual speech recognition systems performs significantly better in such low-resourced settings. Furthermore, what works even better is to generate a language independent symbolic representation of input speech and create NLU systems for this symbolic representation. This hints that creating SLU systems for even extremely low-resourced settings is likely trace conventional SLU pipelines where we represent input speech symbolically in the form of text and then build NLU blocks on top of this. The symbolic representation of speech used here is the phonetic transcription. We find that using a phonetic transcription based system is significantly better than using speech features for classification for low-resourced settings. We see average improvements of 12.37\% and 13.08\% for binary and four-class classification problems respectively, when averaged over 49 different experimental settings, for Belgian Dutch (Flemish) language.




\section{Related Work}
English has been the most widely studied language for creating SLU systems. Various datasets have been released to aid this development \citep{hemphill1990atis, saade2018spoken, lugosch2019speech,  bastianelli-etal-2020-slurp}. There have been many previous works on creating SLU systems in a two-module split fashion \citep{gorin1997may,mesnil2014using}. A typical SLU pipeline, as shown in Figure \ref{fig:1}, consists of an ASR system that converts input speech to text and an NLU module that processes the input text to understand the user query. As with any system composed of multiple modules, errors that occur in one part of the system propagate through the system. To prevent this, a large amount of recent work has been focused on creating E2E-SLU systems \citep{qian2017exploring, serdyuk2018towards, chen2018spoken}. The caveat with making such systems to work is that they require an even larger amount of task-specific annotated data, which is usually not a luxury available to low-resourced languages.


Apart from English, there are many other spoken dialog datasets available for various languages including French \citep{devillers2004french, saade2018spoken}, Dutch \citep{tessema2013metadata, ons2014self, renkens2014acquisition}, Chinese Mandarin \citep{zhu2019catslu, guo2021word}, Sinhala and Tamil \cite{karunanayake2019transfer}, and cross-lingual SLU datasets exist for English, Spanish and Thai \cite{schuster-etal-2019-cross-lingual}. In this paper, we work with two languages - Belgian Dutch (Flemish) \citep{tessema2013metadata, ons2014self, renkens2014acquisition} and English \cite{lugosch2019speech}.

One of the major bottlenecks in creating SLU systems for low-resourced languages is the creation of ASR systems in such low data scenario. This scenario is what we refer to as a language-specific low-resourced setting. Previous works have tried to use English-based ASR systems for languages like Tamil and Sinhala. In these sytems, input speech in Sinhala/Tamil is converted into English script using an English speech recognition system that is then processed by an NLU system \citep{buddhika2018domain, karunanayake2019transfer, karunanayake2019sinhala}. We use a similar idea as baseline and use Wav2Vec \citep{schneider2019wav2vec, baevski2020wav2vec} to extract speech features for Flemish. Wav2Vec is a self-supervised speech recognition system trained on large amounts of unlabelled speech data which boasts to learn superior language representations for English. In this work, we use Wav2Vec 2.0 \cite{baevski2020wav2vec} to extract speech features. 

A series of recent works \citep{gupta2020mere, gupta2020acoustics, gupta2021intent, yadav2021intent} replace the ASR module in the SLU pipeline by a universal phone recognition system called Allosaurus \cite{li2020universal}.  Allosaurus is a universal phonetic transcription system that creates language and speaker independent representations of input speech. Allosaurus is trained to recognize and transcribe input speech into a series of phones contained in the utterance, providing superior representations of input audio which can also be used for languages linguistically distant from high resourced languages like English. \citep{yadav2021intent} show that using embeddings generated from Allosaurus to encode speech content outperforms previous state-of-the-art methods for Sinhala and Tamil by large margins, while maintaining high performance on high resourced languages like English (99.08\% classification accuracy for a 31-class intent classification problem). But the performance drops as the dataset size decreases and is not optimal for the task-specific low resourced settings that we are dealing with in this paper. To tackle this, we convert input speech into phonetic transcriptions using Allosaurus as proposed in \cite{gupta2020acoustics} for our compounded low resourced setting.


In our paper, we explore a novel and rather unexplored language-specific low-resourced setting compounded with task-specific low-resourced-ness. Our aim it to push the limits and demonstrate performance of using existing technologies in extremely low resourced settings, where each data point becomes crucial. 

\begin{figure*}[ht]
  \centering
    \includegraphics[width=0.80\linewidth]{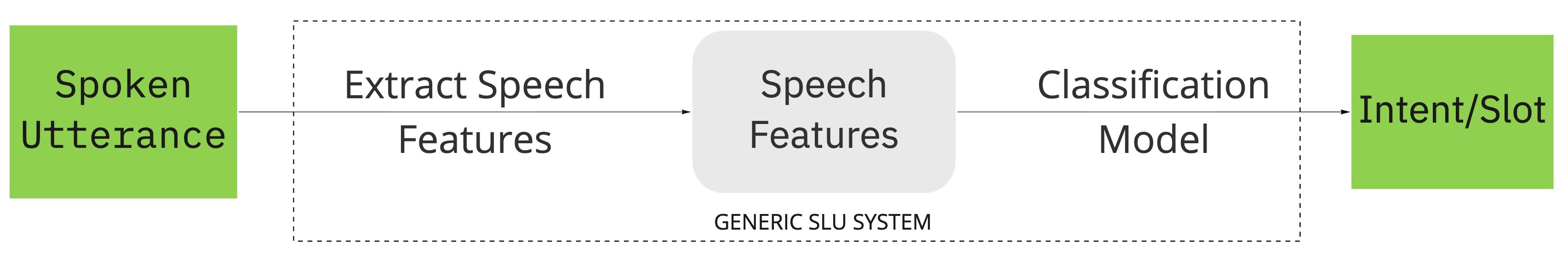}
    \caption{A generic SLU system for language-specific low-resourced setting where we do not have access to speech recognition technologies.}\label{fig:2}
\end{figure*}

\begin{table*}
\centering
\scalebox{0.7}{

\begin{tabular}{c|c|c|c|c|c|c}
\multicolumn{1}{p{2cm}|}{\centering \textbf{Dataset}} & \multicolumn{1}{p{2cm}|}{\centering \textbf{Number of Intents}} & \multicolumn{1}{p{5cm}|}{\centering \textbf{Chosen Intents}} &  \multicolumn{1}{p{2.5cm}|}{\centering \textbf{Speakers in Validation Set}} &  \multicolumn{1}{p{2.5cm}|}{\centering \textbf{Utterances in Validation Set}}&  \multicolumn{1}{p{2cm}|}{\centering \textbf{Speakers in Test Set}} &  \multicolumn{1}{p{2cm}}{\centering \textbf{Utterances in Test Set}}\\
\hline 
FSC (English) & 2 & \multicolumn{1}{p{5cm}|}{\centering `bring newspaper', `activate washroom lights' } & 10 & 194 & 10 & 232    \\

FSC (English) & 4 & \multicolumn{1}{p{5cm}|}{\centering `bring newspaper', `activate washroom lights', `change language to German', `decrease volume'} & 10 & 519 & 10 & 634\\

Grabo (Flemish) & 2 & \multicolumn{1}{p{5cm}|}{\centering approach', `lift' }& 2 & 106 & 2 & 108  \\

Grabo (Flemish) & 4 &  \multicolumn{1}{p{5cm}|}{\centering approach', `lift', `point', `grab'} & 2 & 212 & 2 & 216   \\
\end{tabular}}
\caption{Validation and Test Set statistics for chosen intents for the FSC and Grabo dataset.}\label{Table:Dataset}
\end{table*}

\section{Dataset}\label{sec:dataset}
In our paper, we work with two languages - Belgian Dutch (Flemish) and English.
We use two popular SLU datasets for our experiments - the Fluent Speech Commands (FSC) dataset \cite{lugosch2019speech} for the English language and the Grabo dataset \citep{tessema2013metadata, ons2014self, renkens2014acquisition} for Flemish. 

The primary reason behind the choice of the datasets was that each utterance in the two datasets had clear speaker identities associated with each utterance. Our aim is to test true low resourced settings where getting speaker recordings is extremely hard. Intent recognition datasets in other languages like French \citep{devillers2004french, saade2018spoken}, Chinese Mandarin \citep{zhu2019catslu, guo2021word}, Sinhala and Tamil \cite{karunanayake2019transfer} do not maintain speaker identities and hence were not suitable for our work. Maintaining a mapping of (anonymized) speaker identities allowed us to create validation and test sets with no speaker overlap with the training set. This allows us to do the most robust evaluation of our systems. Moreover, these datasets also allow us to create large test sets such that the results are robust enough to evaluate the system performance and yet have no overlapping speakers with the training set. We choose Flemish as our low-resourced language since Flemish is not used to train Allosaurus or Wav2Vec 2.0. 

FSC is a large and well maintained SLU dataset for the English language. The dataset contains 19 hours of speech data collected from 97 different speakers. The dataset contains commands suitable for a smart home system. An example command would be asking the system to `change language to Chinese' or to `turn off the lights in the kitchen'. Each utterance has a clear, anonymized speaker identity associated with it. This allows us to create large validation and test sets with no speakers overlap with the training set. The intents chosen for our experiments and the corresponding number of samples in the validation and test sets are shown in Table \ref{Table:Dataset}.

The Grabo dataset contains 11 speakers and is much smaller than FSC. The dataset consists of commands given to a robot such as `moving right' or `drive backwards fast'. We use speaker IDs 2-8 to create the training set, speakers 9 and 10 for the validation set, and speakers 11 and 12 for the test set. Thus there is no speaker overlap between the training, validation and test sets. The chosen intents and the validation and test set statistics are shown in Table \ref{Table:Dataset}.

\begin{figure*}[ht]
  \centering
    \includegraphics[width=0.80\linewidth]{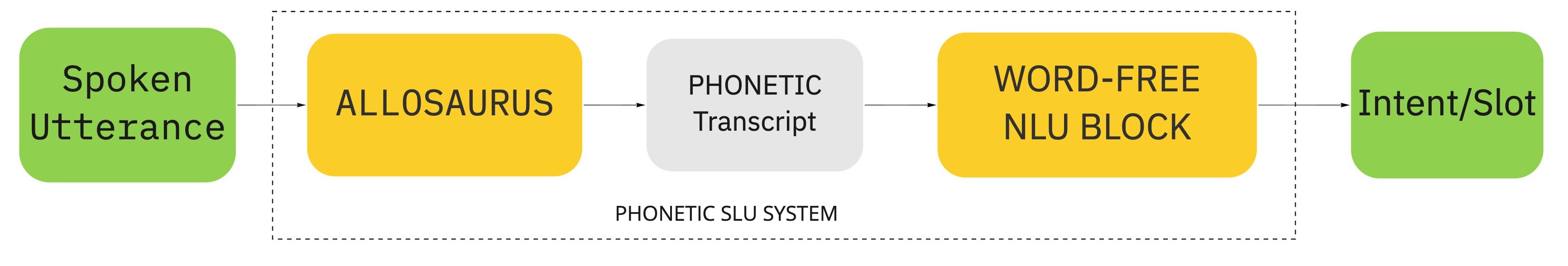}
    \caption{Phonetic transcription based SLU system as proposed in \cite{gupta2020acoustics}.}\label{fig:3}
\end{figure*}

\section{System and Model}\label{sec:model}
To simulate a language-specific low-resourced setting, we do not use a language specific ASR system. We tackle this challenge by exploring two experimental settings. First we use a generic SLU pipeline as shown in Figure \ref{fig:2}. The first step in this pipeline is to extract speech features. We use Wav2Vec 2.0 to extract speech features for Flemish, which represents using a speech recognition system built for a different language. Then, we use the SLU system proposed in \citep{gupta2020acoustics} as shown in Figure \ref{fig:3}. It replaces a language specific ASR system with Allosaurus \cite{li2020universal}, which is a universal phonetic transcription system. We use Allosaurus to convert input speech to phonetic transcriptions. We then build an NLU system from these phonetic transcriptions to perform intent recognition.

The model used in this work is very similar to the model used in \cite{gupta2020acoustics} which is a character level model built for a sequence of phones generated by Allosaurus. The model creates its own embeddings using the annotated task-specific dataset and uses Convolutional Neural Networks (CNN) \cite{lecun1998gradient} to extract contextual information from phonetic input, and a Long-Short Term Memory (LSTM) \cite{hochreiter1997long} network to make utterance level decision and account for sequential information. This model achieved state-of-the-art intent classification performance for low-resourced languages like Tamil and Sinhala when used without language specific ASR. We keep the model used across experiments constant to identify difference in performance occurring due to difference in feature extraction methods.

We reduce the model size to account for the scarcity of data. We use a 256-dimensional embedding layer with just one CNN layer of kernel size 3 and one or two LSTM layers of hidden dimension 256 depending on the dataset size. For the case of the generic SLU, the embeddings are removed and input feature dimension is dependent on the features extracted. For Wav2Vec 2.0, the feature dimensions are 768. A detailed description of model architecture is provided in the appendix \ref{sec:appendix}. Batch normalization \cite{ioffe2015batch} layer is removed because there are scenarios where we are working with a training set of as low as 2 samples, which are not enough to learn batch statistics and give unstable performance.

\section{Experiments}
In this paper, we try to emulate a real world low-resourced data collection scenario. A challenging aspect of building SLU systems for low resourced languages is having access to language specific ASR systems. To tackle this, we experiment with two alternatives. We first use a speech recognition systems created for a higher resourced language (English) to extract speech features and use those features for intent recognition on Flemish data (Section \ref{exp:wav2vec}). Then, we create an intent recognition system using a phonetic transcription generated by Allosaurus (Section \ref{exp:allo}). The input audio is converted to language independent phonetic transcriptions, and intent classification is done using the phonetic transcriptions generated. 


Data collection is expensive and difficult, even more so in extremely low resourced languages. For example, Canadian Indigenious languages like Inuktitut or Siksika have only a few thousand living speakers. Native speakers of such languages are hard to catch hold of for data collection process. This makes every data point collected crucial. This task-specific low-resourced setting compounds the difficulty in making speech technologies for low-resourced languages.

We pose two $I$-class intent classification problems, where $I=2,4$. The columns of each of the Tables \ref{Table:w2vGrabo2}-\ref{Table:Eng2} in the following sections show results for different values of $k$, where $k$ is the number of utterances recorded by a speaker per intent. This means that if $k=3$, each speaker provided $3$ recordings for each intent, which amounts to a total of $3*I$ recordings per speaker. In general, each speaker records $k * I$ audios, where $k$ is the number of audios recorded by a speaker per intent, and $I$ is the number of intents. The rows for each of the tables represent the number of speakers (S) involved in creating the dataset. The total training dataset size is $ S * k * I$. All data points in all the following tables represent an average classification accuracy over 3 different random selections of dataset and training the model from scratch on top of it.


\subsection{Experiments with Wav2Vec Features}\label{exp:wav2vec}
First, we use Wav2Vec 2.0 \cite{baevski2020wav2vec} to extract representations of input speech and use those to perform intent classification on Flemish data. The results for the binary classification setting are shown in Table \ref{Table:w2vGrabo2} and for the four-class classification setting is shown in Table \ref{Table:w2vGrabo4}.

One obvious trend to notice here is that increasing the number of total training samples in general increases the accuracy of the models. This trend is consistently seen in the four-class classification results ( Table \ref{Table:w2vGrabo4}). We also notice a saturation in performance on increasing the number of utterances per speaker. This usually occurs around $k = {4,5}$. For each value of $S$, we see that adding number of recordings for the same speaker increases the performance significantly, but the rate of this increase starts to reduce when we have $4-5$ utterances per speaker. 



\subsection{Experiments with Phonetic Transcriptions using Allosaurus}\label{exp:allo}

The performance in the compounded low-resourced intent classification setting using Wav2Vec features as seen in the previous was encouraging. In this section, we use Allosaurus to generate phonetic transcriptions of user audio, using the pipeline shown in Figure \ref{fig:3}. We then build intent classification systems on top of these phonetic transcriptions. The results for the binary classification setting are shown in Table \ref{Table:Dutch1} and for the four-class classification setting in Table \ref{Table:Dutch2}. 

We consistently see better classification performances for almost all experiments when using phonetic transcriptions. We see an average improvement of 12.37\% for the binary classification problem and 13.08\% for the four-class classification problem, when averaged over 49 different experiments performed in each I-class classification problem. Each experiment represents a accuracy averaged over 3 different random selections of the dataset. Note that the test sets in all the experiments for the binary classification problem are exactly the same with no speaker overlap with the training or the validation set, irrespective of the size of the training set. The same is true for the four-class classification problem.

\begin{table}
\centering

\scalebox{0.7}{

\begin{tabular}{c||l|l|l|l|l|l|l}
    & \textbf{$k=1$} & \textbf{$k=2$} & \textbf{$k=3$} & \textbf{$k=4$} & \textbf{$k=5$} & \textbf{$k=6$} & \textbf{$k=7$} \\             \cline{1-8}
 
  $S = 1$ & 72.53 & 74.69 & 69.44 & 72.83 & 74.07 & 74.38 & 74.07 \\      \hline
  $S = 2$ & 69.75 & 74.69 & 67.90 & 63.27 & 78.70 & 67.59 & 69.13\\      \hline
  $S = 3$ & 68.20 & 76.85 & 82.40 & 80.86 & 76.85 & 74.38 & 72.83\\      \hline
  $S = 4$ & 78.39 & 64.50 & 69.13 & 71.60 & 75.92 & 76.85 & 75.30 \\      \hline
  $S = 5$ & 70.98 & 74.07 & 75.92 & 78.39 & 82.09 & 78.70 & 76.23 \\      \hline
  $S = 6$ & 79.62 & 75.61 & 87.03 & 83.95 & 84.56 & 83.33 & 93.82 \\      \hline
  $S = 7$ & 75.00 & 76.85 & 89.19 & 85.49 & 91.66 & 91.97 & 94.44  \\      \hline
            
\end{tabular}}

\caption{\label{Table:w2vGrabo2}
Two class classification results for the Grabo dataset with 768 dimensional features from Wav2Vec 2.0.}
\end{table} 

\begin{table}
\centering

\scalebox{0.7}{

\begin{tabular}{c||l|l|l|l|l|l|l}
    & \textbf{$k=1$} & \textbf{$k=2$} & \textbf{$k=3$} & \textbf{$k=4$} & \textbf{$k=5$} & \textbf{$k=6$} & \textbf{$k=7$}  \\             \cline{1-8}
 
  $S = 1$ & 35.49 & 37.03 & 37.19 & 37.80 & 39.96 & 40.12 & 42.28\\      \hline
  $S = 2$ & 39.19 & 45.21 & 45.21 & 45.83 & 48.76 & 49.69 & 53.08\\      \hline
  $S = 3$ & 41.82 & 47.83 & 53.70 & 61.57 & 55.55 & 63.88 & 67.59\\      \hline
  $S = 4$ & 49.22 & 45.06 & 51.23 & 52.93 & 60.80 & 65.27 & 64.50\\      \hline
  $S = 5$ & 44.59 & 53.39 & 56.32 & 66.04 & 64.96 & 70.83 & 66.82\\      \hline
  $S = 6$ & 48.14 & 52.77 & 58.64 & 71.91 & 74.07 & 74.69 & 75.30\\      \hline
  $S = 7$ & 52.77 & 56.66 & 67.12 & 72.83 & 79.62 & 80.09 & 76.69\\      \hline
            
\end{tabular}}

\caption{\label{Table:w2vGrabo4}
Four class classification results for the Grabo datasetwith 768 dimensional features from Wav2Vec 2.0.
}
\end{table}

For the binary classification in Flemish, we see that the improvement in performance when using phonetic transcription becomes more significant as the dataset size reduces. This can be observed when we look at the first 3 columns of Table \ref{Table:Dutch1} when compared to Table \ref{Table:w2vGrabo2}. For example, when $S = 7$ and $k \in [5,7]$, the performance of the Wav2Vec system is comparable to the phonetic transcription based system. In all other experiments, the phonetic transcription based system outperforms the Wav2Vec feature based system. Table \ref{Table:Dutch1} also shows that using just 2-3 speakers are enough to learn generalizable speaker independent features when using Allosaurus phonetic transcription, which allows the classification performance on the test set to be in the 90's. A similar performance requires 6-7 speakers when using Wav2Vec features as shown in Table \ref{Table:w2vGrabo2}. This can be seen if we look at a system developed with 3 speakers recording 4 utterances each using phonetic transcriptions in Table \ref{Table:Dutch1}, it is comparable to a 7 speaker system where each speaker records 7 utterances per intent when using Wav2Vec features (Table \ref{Table:w2vGrabo2}) . We attribute this effect to Allosaurus that creates speaker independent embeddings of input audio. These embeddings when projected to the space of a universal set of phones is more robust to speaker variations.

\begin{table}
\centering

\scalebox{0.7}{

\begin{tabular}{c||l|l|l|l|l|l|l}
    & \textbf{$k=1$} & \textbf{$k=2$} & \textbf{$k=3$} & \textbf{$k=4$} & \textbf{$k=5$} & \textbf{$k=6$} & \textbf{$k=7$} \\             \cline{1-8}
 
  $S = 1$ & 75.30 & 81.17 & 73.45 & 79.93 & 76.23 & 82.40 & 78.39 \\      \hline
  $S = 2$ & 84.87 & 85.49 & 93.82 & 89.81 & 87.65 & 91.35 & 89.50\\      \hline
  $S = 3$ & 79.94 & 95.37 & 87.65 & 92.90 & 90.12 & 94.75 & 92.59\\      \hline
  $S = 4$ & 83.33 & 90.74 & 93.20 & 95.06 & 88.58 & 95.37 & 92.28 \\      \hline
  $S = 5$ & 86.11 & 92.59 & 92.90 & 91.35 & 96.29 & 94.75 & 97.83\\      \hline
  $S = 6$ & 91.04 & 91.97 & 92.28 & 94.13 & 96.91 & 91.97 & 92.28 \\      \hline
  $S = 7$ & 85.80 & 90.74 & 90.74 & 90.43 & 94.44 & 91.66 & 95.06  \\      \hline
            
\end{tabular}}

\caption{\label{Table:Dutch1}
Two class classification results for the GRABO (Flemish) dataset using phonetic transcriptions. 
}
\end{table}

\begin{table}
\centering

\scalebox{0.7}{

\begin{tabular}{c||l|l|l|l|l|l|l}
    & \textbf{$k=1$} & \textbf{$k=2$} & \textbf{$k=3$} & \textbf{$k=4$} & \textbf{$k=5$} & \textbf{$k=6$} & \textbf{$k=7$}  \\             \cline{1-8}
 
  $S = 1$ & 47.83 & 50.61 & 50.92 & 53.85 & 50.77 & 52.31 & 50.00\\      \hline
  $S = 2$ & 56.48 & 64.50 & 66.82 & 65.89 & 67.74 & 72.22 & 68.82\\      \hline
  $S = 3$ & 59.87 & 63.58 & 68.36 & 69.90 & 69.75 & 72.22 & 70.52\\      \hline
  $S = 4$ & 63.88 & 64.19 & 68.36 & 67.43 & 72.22 & 71.75 & 73.76\\      \hline
  $S = 5$ & 64.66 & 67.28 & 69.44 & 74.84 & 72.22 & 77.31 & 76.69\\      \hline
  $S = 6$ & 66.51 & 69.59 & 77.93 & 77.46 & 79.62 & 80.55 & 82.56\\      \hline
  $S = 7$ & 68.51 & 80.55 & 81.01 & 82.09 & 85.33 & 85.64 & 88.73\\      \hline
            
\end{tabular}}

\caption{\label{Table:Dutch2}
Four class classification results for the GRABO (Flemish) dataset using phonetic transcriptions. 
}
\end{table}

The performance improvement observed for Flemish when using phonetic transcriptions gets amplified in the four-class classification problem. We see significant improvements when using phonetic transcriptions for all experiments. We see an average improvement of 13.08\% over the 49 experiments when using phonetic transcriptions. This improvement is large when the amount of data is small which we can check by comparing the first three columns of Tables \ref{Table:w2vGrabo4} and \ref{Table:Dutch2}. If we calculate the improvement when $S \leq 3$ and $ k \leq 3$, which we call the $3 \times 3$ matrix of the tables, we get an average improvement of 16.25\% over the 9 experimental settings. But we also see significant improvement when the amount of data is larger. For example, phonetic transcription based system performs significant better for 7 speakers and 7 recording per speaker when compared to the Wav2Vec features based system. Thus, as the task complexity increases, we see that using phonetic transcriptions is a significantly better option when compared to features from speech-to-text systems created for a different language. 

\begin{table}
\centering

\scalebox{0.7}{

\begin{tabular}{c||l|l|l|l|l|l|l}
    & \textbf{$k=1$} & \textbf{$k=2$} & \textbf{$k=3$} & \textbf{$k=4$} & \textbf{$k=5$} & \textbf{$k=6$} & \textbf{$k=7$} \\             \cline{1-8}
 
  $S = 1$ & 72.84 & 82.32 & 84.48 & 86.20 & 79.45 & 83.90 & 86.78 \\      \hline
  $S = 2$ & 84.05 & 89.79 & 91.23 & 86.20 & 94.10 & 94.10 & 95.11\\      \hline
  $S = 3$ & 77.29 & 87.78 & 93.82 & 95.40 & 98.27 & 96.55 & 97.98\\      \hline
  $S = 4$ & 84.33 & 89.51 & 93.10 & 94.97 & 98.41 & 98.85 & 98.13 \\      \hline
  $S = 5$ & 86.20 & 89.65 & 95.25 & 97.27 & 98.13 & 98.70 & 98.27\\      \hline
  $S = 6$ & 86.06 & 95.25 & 96.55 & 98.56 & 98.70 & 97.70 & 99.13 \\      \hline
  $S = 7$ & 96.69 & 95.97 & 96.26 & 98.70 & 99.13 & 98.85 & 98.85  \\      \hline
            
\end{tabular}}

\caption{\label{Table:wv2Eng2}
Two class classification results for the FSC (English) Dataset using speech features extracted from Wav2Vec 2.0.
}
\end{table} 

\begin{table}
\centering

\scalebox{0.7}{

\begin{tabular}{c||l|l|l|l|l|l|l}
    & \textbf{$k=1$} & \textbf{$k=2$} & \textbf{$k=3$} & \textbf{$k=4$} & \textbf{$k=5$} & \textbf{$k=6$} & \textbf{$k=7$}  \\             \cline{1-8}
 
  $S = 1$ & 38.53 & 42.79 & 50.36 & 58.41 & 59.20 & 56.15 & 62.19\\      \hline
  $S = 2$ & 46.58 & 53.73 & 62.56 & 64.30 & 75.23 & 77.97 & 84.01\\      \hline
  $S = 3$ & 48.63 & 58.25 & 75.44 & 85.80 & 81.65 & 81.80 & 92.74\\      \hline
  $S = 4$ & 51.84 & 76.39 & 77.70 & 87.22 & 89.53 & 94.00 & 96.89\\      \hline
  $S = 5$ & 77.86 & 81.59 & 86.33 & 91.48 & 95.58 & 96.79 & 96.31\\      \hline
  $S = 6$ & 72.02 & 90.37 & 81.75 & 95.58 & 95.58 & 95.58 & 97.05\\      \hline
  $S = 7$ & 65.87 & 85.06 & 92.32 & 94.21 & 95.26 & 97.21 & 94.79\\      \hline
            
\end{tabular}}

\caption{\label{Table:w2vEng4}
Four class classification results for the FSC (English) Dataset using speech features extracted from Wav2Vec 2.0.
}
\end{table}

The pipeline proposed in Figure \ref{fig:3} is analogous to the traditional SLU pipline as shown in \ref{fig:1}. High resourced languages allows the use of ASR systems which project speech, which is a very long sequence of high dimensional input into a much shorter, 1-dimensional sequence of characters. Thus, ASR systems try to give a 1-dimensional symbolic representation to input speech. This sequence of characters is usually grouped into words or sub-words, which we refer to as tokens in general, and are then projected back into a higher dimensional space as word-embeddings, encoding meaning and context. This is usually done using pre-trained models like BERT \cite{devlin2018bert}, where the different layers of the model encode and understand various possible meanings and contexts in which a token can be used \cite{tenney-etal-2019-bert}. Thus, these pre-trained models can be seen as functions that map an input token into vectors that encode all possible ways the token has been used in the dataset the model is trained on. 

The projection by ASR systems into a lower dimensional space of characters causes loss of information and results in errors which is not always compensated by the re-projection of words into the space of word-embeddings, which is why recent research in high resourced languages is moving towards creating E2E models. But this process of projecting high-dimensional and long speech input into a much smaller transcription of symbols, and then re-projecting into the space of word-embeddings encoding meaning and context allows us to create SLU systems with a very small amount of annotated task-specific data. 

\begin{table}
\centering

\scalebox{0.7}{

\begin{tabular}{c||l|l|l|l|l|l|l}
    & \textbf{$k=1$} & \textbf{$k=2$} & \textbf{$k=3$} & \textbf{$k=4$} & \textbf{$k=5$} & \textbf{$k=6$} & \textbf{$k=7$} \\             \cline{1-8}
 
  $S = 1$ & 91.98 & 93.27 & 95.56 & 95.27 & 95.85 & 96.71 & 96.56 \\      \hline
  $S = 2$ & 95.13 & 97.99 & 97.99 & 98.56 & 98.56 & 98.14 & 97.28\\      \hline
  $S = 3$ & 95.85 & 98.28 & 97.85 & 97.65 & 99.14 & 99.71 & 99.28\\      \hline
  $S = 4$ & 97.28 & 98.42 & 98.14 & 98.88 & 98.99 & 98.85 & 98.71 \\      \hline
  $S = 5$ & 98.56 & 97.56 & 98.99 & 98.71 & 99.28 & 98.85 & 99.28\\      \hline
  $S = 6$ & 96.71 & 97.85 & 98.42 & 98.56 & 98.56 & 98.71 & 99.58 \\      \hline
  $S = 7$ & 97.42 & 99.57 & 99.42 & 99.71 & 99.85 & 99.57 & 99.42  \\      \hline
            
\end{tabular}}

\caption{\label{Table:Eng1}
Two class classification results for the FSC (English) using phonetic transcriptions.
}
\end{table} 

\begin{table}
\centering

\scalebox{0.7}{

\begin{tabular}{c||l|l|l|l|l|l|l}
    & \textbf{$k=1$} & \textbf{$k=2$} & \textbf{$k=3$} & \textbf{$k=4$} & \textbf{$k=5$} & \textbf{$k=6$} & \textbf{$k=7$}  \\             \cline{1-8}
 
  $S = 1$ & 61.06 & 62.06 & 62.79 & 70.59 & 69.75 & 72.29 & 72.21\\      \hline
  $S = 2$ & 65.04 & 63.99 & 72.05 & 77.18 & 80.06 & 78.64 & 81.84\\      \hline
  $S = 3$ & 67.13 & 74.72 & 75.35 & 77.91 & 83.72 & 85.55 & 85.29\\      \hline
  $S = 4$ & 68.60 & 79.74 & 77.18 & 84.66 & 84.51 & 88.54 & 87.75\\      \hline
  $S = 5$ & 72.05 & 79.59 & 80.58 & 87.85 & 88.27 & 91.57 & 92.67\\      \hline
  $S = 6$ & 70.80 & 82.20 & 83.41 & 90.16 & 89.84 & 91.05 & 92.83\\      \hline
  $S = 7$ & 75.56 & 80.48 & 86.65 & 89.48 & 91.10 & 90.99 & 93.98\\      \hline
            
\end{tabular}}

\caption{\label{Table:Eng2}
Four class classification results for the FSC (English) dataset using phonetic transcriptions.
}
\end{table} 

Our experiments show that the analogous process of projecting down speech into a symbolic transcription of phones and then re-projecting the symbols into a vector space of symbolic embeddings created from the phonetic transcription data performs significantly better than using high dimensional feature representations of input speech, as done with Wav2Vec in section \ref{exp:wav2vec}. The large size of Wav2Vec vectors (768) requires a larger amount of task-specific data to infer content and meaning of input utterances when compared to using phonetic transcription. Using phonetic transcriptions also allow us to create our own vector spaces of symbolic embeddings which are very specific to our dataset and encode the meaning and context in which each phone has been used for the particular task. This is why the pipeline that uses phonetic transcriptions outperforms Wav2Vec based embeddings. \cite{yadav2021intent} show that this is true even when Allosaurus embeddings are compared to phonetic transcriptions generated by Allosaurus. As the amount of available data decreases, intent classification systems built using phonetic transcriptions begin to outperform systems based on Allosaurus embeddings, thus showing that projecting input speech into phonetic transcriptions is the most exhaustive way to use the scarce amount of labelled data in the compounded low-resourced settings.

We verify this by performing the same set of experiment on the English dataset (FSC). We first use Wav2Vec features to extract input speech. The binary classification, the results are shown in Table \ref{Table:wv2Eng2} and for the four-class classification problem, the results are shown in Table \ref{Table:w2vEng4}. Note that Wav2Vec is specifically trained on large amounts of English speech data and thus the features extracted from Wav2Vec are likely to perform much better for English than they worked for Flemish. This experimental setting is thus not a language-specific low-resourced setting anymore, and only a task-specific low-resourced setting.  We then create an intent classification system using phonetic transcriptions, as shown in Table \ref{Table:Eng1} and \ref{Table:Eng2}. We see an average improvement of 5.42\% for the binary classification problem and 2.09\% for the four-class classification problem, when averaged over 49 experiments. These improvements are amplified when we compare the $3 \times 3$ matrices (when $S \leq 3$ and $ k \leq 3$, ) for the two classification problems between Wav2Vec based and phonetic transcription based methods. We find an average improvement of 11.14\% for the binary classification problem and an average improvement of 14.15\% for the four-class classification problem, when averaged over 9 experiments. This shows that a phonetic transcription based SLU pipeline outperforms a speech feature-based pipeline in the low-resourced scenarios, especially when we lack language specific speech recognition technologies.

\section{Conclusion}

In this paper, we provide a series of experiments to empirically recreate a real-world, low-resourced, SLU system building scenario. We work in the compounded setting of language-specific low-resourced-ness and task-specific low-resourced-ness. The challenge posed by a language-specific low-resourced setting is the absence speech recognition technologies. We bypass this in two ways - firstly, we use a speech recognition system built for a different higher resourced language. Secondly, we use a universal phone recognition system to convert input speech to phonetic transcriptions. To simulate the task-specific low-resource scenario, we present intent classification results at a granularity where we see the effects of changing the number of speakers and the utterances recorded by each speaker. We simulate these settings for Belgian Dutch (Flemish) and English.

We find that using Allosaurus, a universal phone recognition system that creates language and speaker independent representations of input speech, performs better than using Wav2Vec for Flemish dataset. When using Allosaurus, we convert input speech into phonetic transcriptions and use these transcriptions to build NLU models. We find that using phonetic transcription based model performs better than using Wav2Vec features. For Flemish, we see an average improvement of 12.37\% for a binary classification problem and an average improvement of 13.08\% for a four-class classification over using Wav2Vec features, when averaged over 49 different experimental settings. All results are calculated on a large test set containing hundreds of utterances that has no speaker overlap with the training or validation set. Also, we find that as the dataset size decreases, phonetic transcription based method consistently outperform Wav2Vec feature based methods. Phonetic transcription based models also need fewer speakers to generalize to a test set with no speaker overlap.

Finally, we recommend converting input speech into phonetic transcriptions as an intermediate step for creating SLU systems in such low resourced settings. Doing such conversion allows us to create a task-specific embedding space that uses the small annotated dataset most efficiently.

\textbf{\textit{Disclaimer}}. This paper was prepared for informational purposes by the Artificial Intelligence Research group of JPMorgan Chase \& Co and its affiliates (“JP Morgan”), and is not a product of the Research Department of JP Morgan. JP Morgan makes no representation and warranty whatsoever and disclaims all liability, for the completeness, accuracy or reliability of the information contained herein. This document is not intended as investment research or investment advice, or a recommendation, offer or solicitation for the purchase or sale of any security, financial instrument, financial product or service, or to be used in any way for evaluating the merits of participating in any transaction, and shall not constitute a solicitation under any jurisdiction or to any person, if such solicitation under such jurisdiction or to such person would be unlawful. © 2022 JPMorgan Chase \& Co. All rights reserved.

\bibliography{anthology,custom}
\bibliographystyle{acl_natbib}


\appendix

\section{Implementation Details}
\label{sec:appendix}
All models are trained using the NVIDIA GeForce GTX 1070 GPU using python3.7. The training is very quick due to the small dataset sizes, with each epoch taking 1-2 seconds. For each experiment, a validation set identical to the test set was used. For the FSC dataset, the validation set had 10 speakers with no speaker overlap with the training or the test set. Similarly for the GRABO dataset, the validation set had 2 speakers that were not present in the training or the test set. Each experiment in Tables \ref{Table:w2vGrabo2}-\ref{Table:Eng2} was repeated 3 times with a different training set and the average accuracy has been reported. 

As mentioned in section \ref{sec:model}, we use a CNN+LSTM architecture, as proposed in \cite{gupta2020acoustics}. We performed a grid search over various parameters of the architecture. The best performing models varied slightly for each experiment. The exact model parameters for the results reported in Tables \ref{Table:w2vGrabo2}-\ref{Table:Eng2} are shown in Table \ref{Table:Model}. For larger amounts of utterances recorded per speaker, we found better results with 2 LSTM layers instead of one. 

\begin{table}
\centering
\scalebox{0.9}{

\begin{tabular}{c|c}
\multicolumn{1}{p{5cm}|}{\centering \textbf{Model Parameters}} & \multicolumn{1}{|p{3cm}}{\centering \textbf{Value}} \\
\hline 
Embedding Size & 256 \\
CNN kernel size & 3 \\
No. of CNN filters & 256 \\
No. of LSTM layers & 1 ( or 2) \\
LSTM hidden size & 256 \\
Batch Normalization & False
\end{tabular}}
\caption{Model Parameters}\label{Table:Model}
\end{table}

\end{document}